\documentclass[runningheads]{llncs}
\usepackage{graphicx}

\usepackage{tikz}
\usepackage{comment}
\usepackage{amsmath,amssymb} 
\usepackage{color}
\usepackage{multirow}
\usepackage{booktabs}
\usepackage{subfigure}
\usepackage{bbding}

\usepackage[accsupp]{axessibility}  

\begin{document}
\pagestyle{headings}
\mainmatter
\def\ECCVSubNumber{3005}  

\title{Worst Case Matters for Few-Shot Recognition} 

\titlerunning{Worst Case Matters for Few-Shot Recognition}
%
\author{Minghao Fu \and
Yun-Hao Cao \and Jianxin Wu\thanks{J. Wu is the corresponding author.}
}
\authorrunning{M. Fu et al.}
%
\institute{State Key Laboratory for Novel Software Technology\\ 
Nanjing University, Nanjing, China \\
\email\{fumh, caoyh\}@lamda.nju.edu.cn, \ \email wujx2001@gmail.com}

\maketitle

\begin{abstract}
Few-shot recognition learns a recognition model with very few (e.g., 1 or 5) images per category, and current few-shot learning methods focus on improving the average accuracy over many episodes. We argue that in real-world applications we may often only try one episode instead of many, and hence maximizing the worst-case accuracy is more important than maximizing the average accuracy. We empirically show that a high average accuracy not necessarily means a high worst-case accuracy. Since this objective is not accessible, we propose to reduce the standard deviation and increase the average accuracy simultaneously. In turn, we devise two strategies from the bias-variance tradeoff perspective to implicitly reach this goal: a simple yet effective stability regularization (SR) loss together with model ensemble to reduce variance during fine-tuning, and an adaptability calibration mechanism to reduce the bias. Extensive experiments on benchmark datasets demonstrate the effectiveness of the proposed strategies, which outperforms current state-of-the-art methods with a significant margin in terms of not only average, but also worst-case accuracy. Our code is available at \url{https://github.com/heekhero/ACSR}.
\end{abstract}

\section{Introduction}
\label{sec:introduction}

Most people have the ability to learn to recognize new patterns via one or a few samples (e.g., images), thanks to the accumulated knowledge. Naturally, \textit{few-shot learning}~\cite{wang2020generalizing} aims at learning from scarce data, which is already studied long before the deep learning era. In this paper, we focus on the image recognition task, also known as \emph{few-shot image classification}, a widely studied few-shot task~\cite{vinyals2016matching,snell2017prototypical,finn2017model,sung2018learning,chen2019closer,liu2020negative,afrasiyabi2020associative,mangla2020charting,dhillon2020baseline,afrasiyabi2021mixture,yang2021free}. Deep learning techniques have further pushed few-shot learning's \emph{average} accuracy over multiple runs (i.e., \emph{episodes}) towards a high level that appears to be already applicable to real-world applications.

In this task, a pretrained model is first derived from the \emph{base set}, a large set of labeled images. Then, given some unseen categories and very few training images per category (the \emph{novel set}), the model must learn to adapt to classifying new examples from these novel categories. Differently sampled novel sets lead to different episodes, different trained models and test accuracies. The common evaluation criterion is to run a large number of (usually 500 to 10000) episodes for the same task, and report the average accuracy and its 95\% confidence interval.

\begin{figure}[t]
    \centering
    \includegraphics[width=0.57\linewidth]{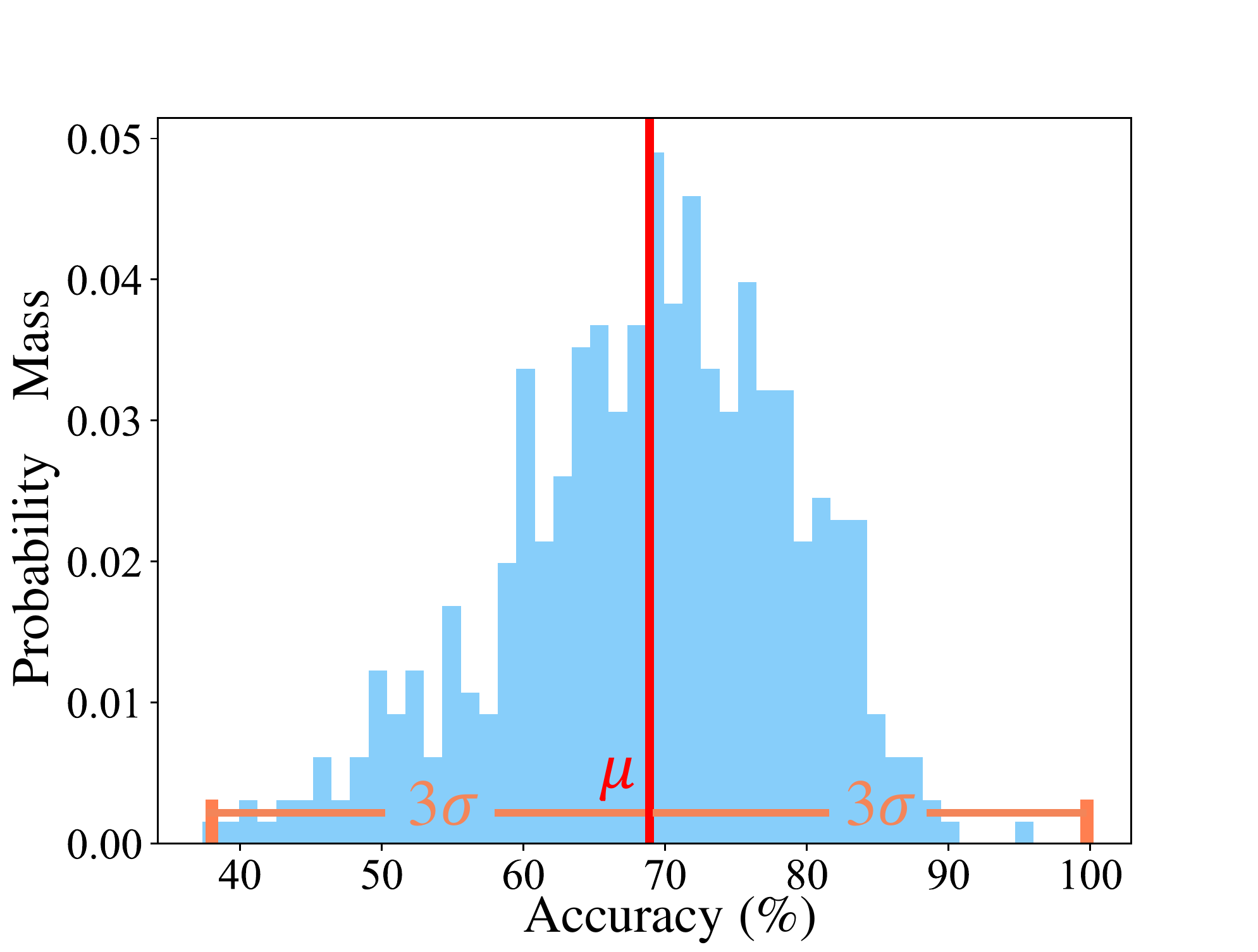}
    \caption{Distribution of accuracy rates of 500 episodes in our experiments, with 5-way 1-shot experiments of the LR-DC~\cite{yang2021free} method on \emph{mini}-ImageNet. Best viewed in color.} 
    \label{fig:hist}
\end{figure}

We aim at making few-shot recognition more practical, too. But, we argue that both metrics (mean accuracy and 95\% confidence interval) are \emph{not helping us towards reaching this goal}, given the recent progress in this task. Fig.~\ref{fig:hist} shows the distribution of accuracy of 500 episodes for the same task, whose average accuracy estimate is 68.96\% and the 95\% confidence interval is $[68.07,69.85]$---this interval has 95\% chance of the true average accuracy landing on it, not the chance of a single episode's accuracy dropping inside! In other words, both metrics are used to describe \emph{the mean accuracy of 500 episodes}, \emph{not the accuracy of a single experiment}.

In fact, the worst episode among these 500 in Fig.~\ref{fig:hist} has only 37.33\% accuracy while that of the best is close to 100\%. That is, few-shot learning is very \emph{unstable}, and the accuracy varies dramatically. The \emph{worst-case lags far behind the average}. 

Furthermore, now that we aim at making few-shot learning practical, we have to accept that in most real-world applications, we can run the experiment \underline{\emph{only once}}, but do not have the luxury of running 500 (or more) experiments and pick the best or average episode among them. In other words, \emph{the worst-case scenario} (if one is unlucky) is naturally more important than the average case---in commercial applications, we welcome a product with an acceptable worst-case performance more than one that performs well on average but occasionally goes calamitous. We argue that we need to pay attention to one episode instead of the average of 500 episodes, and to \emph{maximize the worst case's accuracy} instead of the mean.

But, it is in general very challenging to explicitly optimize the worst-case scenario even when there are many training examples, let alone in the few-shot setting. We propose an alternative, indirect line of attack. As shown in Fig.~\ref{fig:hist}, the accuracy distribution often fits well to a Gaussian. Then, the worst-case accuracy is naturally estimated by the $3\sigma$ rule as $\mu-3\sigma$, where $\mu$ is the average accuracy and $\sigma$ is the accuracy's standard deviation---in a normal distribution, the chance to fall on the left of $\mu-3\sigma$ is only 0.135\%. Fig.~\ref{fig:hist} clearly supports this approximation. In other words, we do need to increase the mean accuracy $\mu$ (as in current methods), but also need to \emph{simultaneously minimize the standard deviation $\sigma$}.

Unfortunately, directly optimizing $\sigma$ is not plausible, because it involves different episodes and different models, which is beyond the capability of end-to-end deep learning. To tackle this issue, we propose to resort to the bias-variance decomposition theory~\cite{hastie2009elements}, which states that the expected error ($1-\mu$) is the sum of squared bias plus variance ($\sigma^2$). From this classic theory, we know that \emph{stable} learning machines lead to small $\sigma$, and hence propose a novel \emph{stability regularization} (SR) to utilize the base set to achieve stability. We also incorporate model ensemble, because it is well-known for its effect of reducing both bias and variance. Similarly, this theory also states that a model with larger capacity often lead to smaller bias, hence we propose an \emph{adaptability calibration} (AC) strategy to properly increase the model capacity without overfitting, so as to indirectly reduce the bias and to increase the average accuracy $\mu$ in turn.

Our contributions can be summarized as follows:
\begin{enumerate}
    \item To the best of our knowledge, we are the first to emphasize the importance and to advocate the adoption of worst case accuracy in few-shot learning.
    
    \item Motivated by Fig.~\ref{fig:hist}, we argue that in addition to maximizing the average accuracy $\mu$, we must also \emph{simultaneously} reduce the standard deviation $\sigma$.

    \item We propose to achieve this goal from the bias-variance tradeoff perspective. We propose a simple yet effective stability regularization (SR) loss together with model ensemble to reduce variance during fine-tuning. The SR loss is computed in an unsupervised fashion, and can be easily generalized to data beyond the base set. We also propose adaptability calibration (AC) to vary the number of learnable parameters to reduce bias.
\end{enumerate}

As a result, our method not only achieves higher average accuracy than current state-of-the-art methods, but more importantly enjoys significantly higher worst case accuracy on benchmark few-shot recognition datasets. 

\section{Related Work}

Generalizing from base categories, few-shot learning aims at designing effective recognition systems with limited training data from novel categories. Current few-shot learning research can be roughly divided into two branches: those based on meta learning and those based on transfer learning. We now give a brief introduction of representative methods within both branches. Besides, since we propose using the ensemble technique for better regularizing the model stability, model ensemble methods in few-shot problems will be discussed as well.

\textbf{Meta learning.} This line of methods model the training process of few-shot learning by pretending the episodes (which are usually large in quantity) are to be learned in a way of ``learning to learn''. One line of work~\cite{finn2017model,lee2019meta,liu2020ensemble,rusu2019meta,finn2018probabilistic} learns a set of good initial weights to fast adapt to unseen episodes with a limited number of gradient descent steps. Another line of work~\cite{snell2017prototypical,sung2018learning,vinyals2016matching,bertinetto2018meta,kim2019variational} leverages the characteristics of different distance metrics to classify unknown samples by comparing with embeddings or their variants (e.g., prototypes) derived from the training examples.

\textbf{Transfer learning.} These methods~\cite{chen2019closer,gidaris2019generating,bateni2020improved,gidaris2019boosting,dvornik2019diversity,liu2020negative,afrasiyabi2020associative,mangla2020charting,dhillon2020baseline,afrasiyabi2021mixture,hu2021leveraging,yang2021free} take advantage of the standard transfer learning pipeline, which first pretrain a model on base classes, then revise the feature embeddings output by the pretrained model with limited novel samples.~\cite{chen2019closer} used cosine classifier to normalize the magnitude of both embeddings and classification weights for compacting the intra-class intensity.~\cite{gidaris2019boosting} proposed to pretrain with self-supervised auxiliary tasks for boosting few-shot learning.~\cite{liu2020negative} introduced a negative margin loss during pretraining to increase the discriminability of features.~\cite{afrasiyabi2020associative} used the associative alignment strategy to learn novel features by aligning them with closely related samples from the base set.~\cite{mangla2020charting} explored the way to pretrain with manifold mixup~\cite{verma2019manifold} for good generalization.~\cite{dhillon2020baseline} proposed a baseline for transductive fine-tuning with novel samples.~\cite{hu2021leveraging} introduced a new pipeline, which first preprocessed features, then classified them with an optimal-transport inspired algorithm.~\cite{yang2021free} adjusted novel distributions using base categories to generate imaginary training samples.

\textbf{Model ensemble.} It is a well-known strategy~\cite{hastie2009elements} to incorporate a number of weak models to build a strong model, which will effectively increase the stability and robustness during inference. Very recently, for the few-shot classification problem, \cite{dvornik2019diversity} harmonized the cooperation and diversity between deep models to pursue better ensemble performance. E$^3$BM~\cite{liu2020ensemble} tackled few-shot problems with the ensemble of epoch-wise base-learners whose hyper-parameters were generated by task-specific information. Different from these methods, our ensemble stability regularization directly increases the diversity of deep models by dealing with different parts of data from the base set, keeping simplicity while achieving superior performance.

\section{The Worst-case Accuracy and Its surrogate}

We first give some background information about how a few-shot problem is defined, then describe the relationship between the commonly used 95\% confidence interval and the standard deviation ($\sigma$). Lastly, our solution that advocates indirectly optimizing for the worst-case will be presented.

In the few-shot recognition setting, there exists a dataset with abundant labeled images called the base set, denoted as $D_b=\{x_i^b, y_i^b \}_{i=1}^{N_b}$, where $x_i^b \in R^D$ is the $i$-th training image, $y_i^b \in \mathcal Y_b$ is its corresponding category label, and $N_b$ is the number of examples. In addition, there also exists another dataset with scarce labeled images from new categories (usually 1 or 5 per category) called the novel set, denoted as $D_n=\{x_j^n, y_j^n\}_{j=1}^{N_n}$, where $y_j^n \in \mathcal Y_n$ and $\mathcal Y_b \cap \mathcal Y_n = \emptyset$, with $N_n$ being the number of examples it contains. In every episode, $N$ categories will be sampled to constitute the novel set, and $K$ examples will be sampled from each of those categories (called $N$-way $K$-shot). Learned with this tiny set of images (and optionally $D_b$), a few-shot recognizer needs to recognize unseen examples from the $N$ categories in this episode.

\subsection{Existing and Proposed Metrics}
\label{sec:conf_2_std}

Existing methods evaluate their performance by averaging the accuracy of $n$ episodes, including the average accuracy $\mu$ and a 95\% confidence interval $[\mu-Z_{95\%},\mu+Z_{95\%}]$. As aforementioned, both metrics are estimates for the average accuracy random variable, not estimating the accuracy of one episode. Hence, the interval radius $Z_{95\%}$ is often surprisingly small, as shown in Table~\ref{tab:pre_results}.

\begin{table*}[t]
\footnotesize
\centering
\caption{5-way 1-shot recognition results of existing methods on \emph{mini}-ImageNet (500 episodes). $\flat$ means that $\sigma$ is calculated by our implementation due to undisclosed $n$ in published papers. $\text{ACC}_{m}$, $\text{ACC}_{1}$ and $\text{ACC}_{10}$: higher is better; $Z_{95\%}$ and $\sigma$: lower is better. The best results are shown in boldface.}  
\resizebox{.6\linewidth}{!}{
    \begin{tabular}{l|ccccc|ccc}
        \toprule
        Method & $\text{ACC}_m$ && $Z_{95\%}$ && $\sigma$ & $\text{ACC}_{1}$ && $\text{ACC}_{10}$ \\
        \midrule
        Negative-Cosine~\cite{liu2020negative} & 61.72 && 0.81 && 10.12 & 24.27 && 36.13 \\
        MixtFSL~\cite{afrasiyabi2021mixture}& 64.31 && 0.79 && 9.87 & 30.67 && 35.07\\
        $\text{S2M2}_R$~\cite{mangla2020charting} & 64.93 && \textbf{0.18} && \textbf{9.18} & 37.58 && 42.87 \\
        PT+NCM~\cite{hu2021leveraging} & 65.35 && 0.20 && 10.20 & 32.00 && 38.13 \\
        CGCS~\cite{Gao_2021_ICCV}& 67.02 && 0.20 && 10.20 & \textbf{38.70} && \textbf{44.00} \\
        LR-DC~\cite{yang2021free} & \textbf{68.57} && 0.55 && 10.28$^\flat$ & 37.33 && 42.72 \\
        \bottomrule
    \end{tabular}}
    \label{tab:pre_results}
\end{table*}

As established in Sec.~\ref{sec:introduction}, we need to focus more on the worst-case accuracy among all $n$ episodes, which is denoted as $\text{ACC}_1$. In addition, we also report the average accuracy of the 10 worst cases as $\text{ACC}_{10}$. The empirical average accuracy is denoted as $\text{ACC}_m$. Although it is a general trend that $\text{ACC}_m$ and $\text{ACC}_1$ are positively correlated, the 6 methods rank significantly differently using $\text{ACC}_m$ and $\text{ACC}_1$. For example, $\text{S2M2}_R$ is 3.64\% lower than LR-DC in terms of $\mu$ ($\text{ACC}_m$), but 0.25\% higher in $\text{ACC}_1$ (worst-case accuracy). That is, although maximizing $\mu$ is useful, it is far from being enough. We argue that for few-shot recognition to be practically usable, we need to maximize $\text{ACC}_1$ instead.

The $Z_{95\%}$ metric is also misleading, as it measures uncertainty in estimating $\mu$. In fact, based on its definition, we have
\begin{equation}
\sigma = Z_{95\%} \cdot \frac{\sqrt{n}}{1.96} \,,
\label{eq:conf_to_std}
\end{equation}
where $\sigma$ is the standard deviation of accuracy across episodes. Hence, a small $Z_{95\%}$ may well be because $n$ is large (e.g., $n=10000$), instead of due to a small $\sigma$, which is illustrated clearly in Table~\ref{tab:pre_results}. For example, MixtFSL has almost 3x larger $Z_{95\%}$ than that of CGCS, but has a smaller $\sigma$ within the pair. And different papers often use different $n$ values, which renders $Z_{95\%}$ difficult to interpret.

Furthermore, the proposed worst-case accuracy $\text{ACC}_1$ is not only semantically more meaningful than $\text{ACC}_m$ ($\mu$), but also more stable. We use $\text{ACC}_k$ to denote the average accuracy of the $k$ worst episodes. Results in Table~\ref{tab:pre_results} exhibits that $\text{ACC}_1$ and $\text{ACC}_{10}$ ranks the methods consistently.

\subsection{Implicitly Optimizing the Worst-Case Accuracy}

As briefly introduced in Sec.~\ref{sec:introduction}, we can only implicitly maximize $\text{ACC}_1$, and we propose to use $\mu-3\sigma$ as a surrogate. As empirically shown in Fig.~\ref{fig:hist}, when the accuracy distributes as a Gaussian, $\mu-3\sigma$ is a perfect surrogate, as it pins to the 0.135-th percentile of $\text{ACC}_1$, because $\Phi(-3)=0.00135$ in which $\Phi$ is the cumulative distribution function of the standard normal distribution $N(0,1)$. Even if the accuracy distribution is highly non-normal (which empirical data suggests otherwise), the one-sided Chebyshev inequality also guarantees that $\mu-3\sigma$ is no larger than the 10th percentile (i.e., no better than the 10\% worst cases), because the inequality states that $\Pr(X \le \mu -k\sigma) \le \frac{1}{1+k^2}$ for any distribution $X$ and any $k>0$, while $k=3$ in our case.

But, this surrogate loss is still a qualitative one instead of a variable that can be directly maximized, because $\sigma$ non-linearly involves all $n$ episodes, including the $n$ models and the $n$ training sets. Hence, we transform our objective to \emph{simultaneously maximize $\mu$ and minimize $\sigma$}, and propose to use the bias-variance tradeoff for achieving both objectives indirectly.

\subsection{The Bias-Variance Tradeoff in the Few-Shot Scenario}

The bias-variance decomposition states that the expectation of error rate (i.e., $1-\mu$) equals the sum of the squared bias and the variance~\cite{hastie2009elements}.\footnote{Here $\mu$ is the population mean, but we also use the same notation for sample mean. They can be easily distinguished by the context.} Although the definitions for bias and variance of a classifier are not unique, the following qualitative properties are commonly agreed upon~\cite{hastie2009elements}:
\begin{enumerate}
	\item A classifier has large variance if it is unstable (small changes in the input cause large changes in the prediction), and vice versa; Hence, we expect \emph{a smaller $\sigma$ (equivalently, variance) if we make the recognizer more stable}.
	\item A classifier with larger capacity (e.g., more learnable weights) in general has a smaller bias, and vice versa. Hence, we prefer \emph{a model with larger capacity} to reduce the bias.
	\item Although minimizing both bias and variance simultaneously amounts to maximizing the expected accuracy, the two terms are often contradictory to each other. Larger models are more prone to overfitting and in turn larger variance, too. Hence, \emph{a properly calibrated capacity increase} is necessary to strike a balance.
\end{enumerate}

\subsection{Reducing Variance: Stability Regularization}

\begin{figure}[t]
    \centering
    \includegraphics[width=0.8\linewidth]{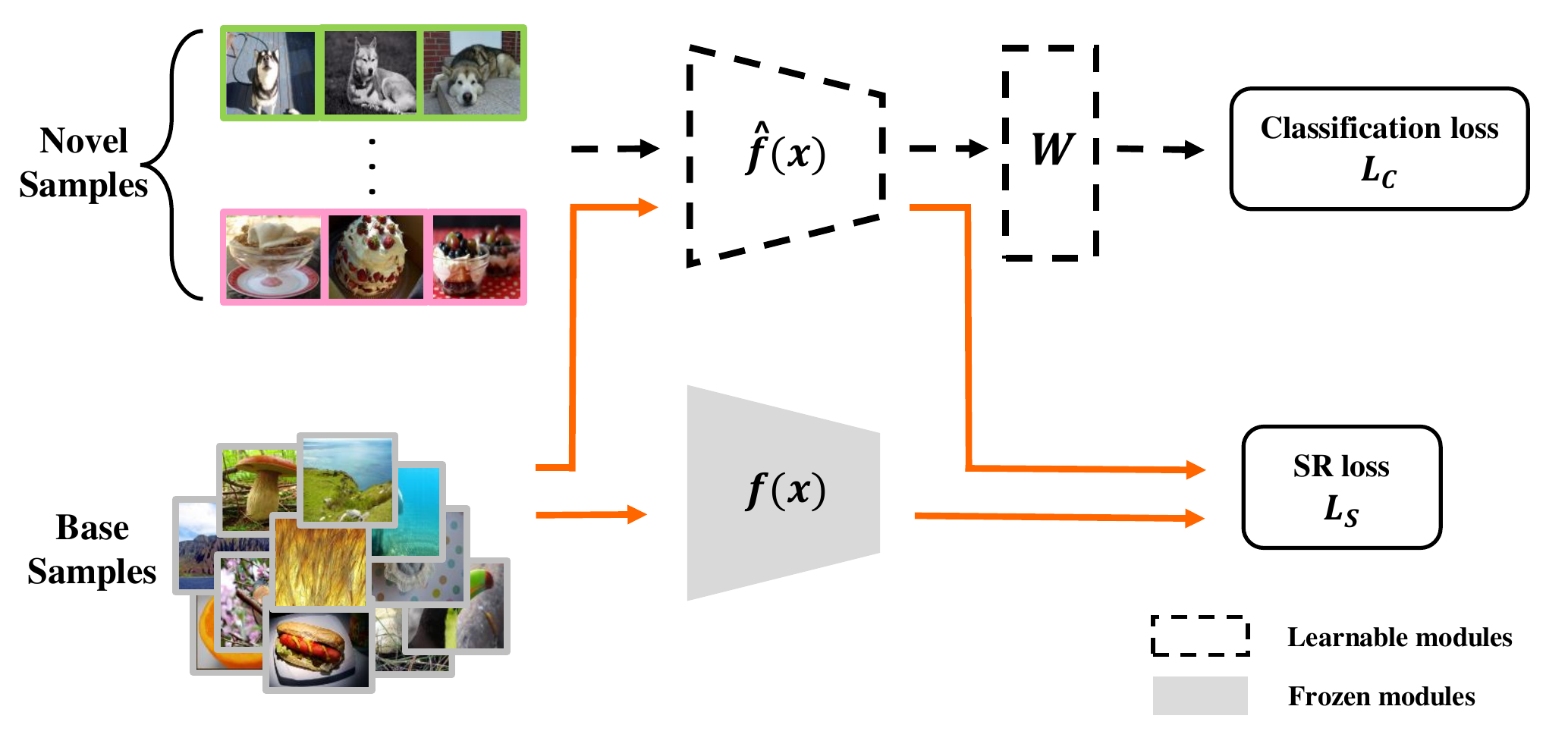}
    \caption{The proposed few-shot recognition framework. A backbone network $f(\mathbf{x})$ is trained using the base set samples and its classification head is discarded. Using images from the novel set, the backbone is fine-tuned into $\hat{f}(\mathbf{x})$, plus a new classification head $W$. Stability of $\hat{f}(\mathbf{x})$ is maintained by the stability regularization (SR) loss $\mathcal{L}_S$, while the capacity increase (adaptability) is calibrated by fine-tuning a selected subset of layers in the backbone $\hat{f}(\mathbf{x})$. The classification loss $\mathcal{L}_C$ minimizes error (i.e., to increase the average accuracy $\mu$). Best viewed in color.}
    \label{fig:network}
\end{figure}

To fulfill these goals, our framework (cf. Fig.~\ref{fig:network}) follows the common practice to train a backbone network $f(\mathbf{x})$ (its classification head discarded) using the base set $D_b$. Then, in one episode, a $N$-way $K$-shot training set is sampled as the novel set $D_n$ to fine-tune the backbone (with a randomly initialized classification head $W$) into $\hat{f}(\mathbf{x})$ using a usual cross entropy classification loss $\mathcal{L}_C$.

The danger is: because $K$ is very small (mostly 1 or 5), the fine-tuning process easily gets overfit and is extremely \emph{unstable}---different sampled training sets lead to dramatically different prediction accuracy. To increase the stability (and thus to reduce the variance $\sigma^2$), we must not allow the weights in $\hat{f}(\mathbf{x})$ to be dominated by the small  novel training set and completely forget the representation learned from the base set, i.e., the knowledge in $f(\mathbf{x})$.

Because the base classification head has been discarded, the stability is regularized by requiring the original learned representation ($f(\mathbf{x})$ for an input $\mathbf{x}$) and the fine-tuned representation ($\hat{f}(\mathbf{x})$ for the same input) to remain similar to each other, or at least not to deviate dramatically. A simple negative cosine similarity loss function realizes this stability requirement:
\begin{equation}
	\mathcal{L}_S(\mathbf{x}) = - \frac{f(\mathbf{x}) \cdot \hat{f}(\mathbf{x})}{\|f(\mathbf{x})\| \|\hat{f}(\mathbf{x})\|} \,.
\end{equation}
In order to avoid overfitting, the $\mathbf{x}$ in our stability regularization (SR) loss is not sampled from the novel set. In each mini-batch, we randomly sample 256 images from the base set with replacement to calculate $\mathcal{L}_S$ and back-propagate to $\hat{f}(\mathbf{x})$ (but $f(\mathbf{x})$ is frozen). The proposed SR loss is minimized if $\hat{f}(\mathbf{x})$ and $f(\mathbf{x})$ are the same (modulo a scale factor), which makes $\hat{f}(\mathbf{x})$ produce similar representations to $f(\mathbf{x})$ for \emph{all} base set images, hence we can expect that it will be stable given different training sets sampled from the novel split.

It is worth noting that the proposed SR loss is very flexible since only unlabeled images are required to compute the loss. Hence, it can be easily extended to use other unlabeled images. Results on using images from other than the base or novel set are presented in Sec.~\ref{sec:ablation}.

The other loss function $\mathcal L_C$ in Fig.~\ref{fig:network} is a regular cross entropy loss, which aims at maximizing the average accuracy $\mu$. Hence, the overall objective is $\mathcal L = \mathcal L_C + \alpha \cdot \mathcal L_S$, where $\alpha$ is always 0.1 in all experiments.

\subsection{Reducing Bias: Adaptability Calibration}

As aforementioned, to reduce the bias, we adjust the capacity of our model. In many few-shot learning methods~\cite{liu2020negative,afrasiyabi2021mixture,yang2021free,snell2017prototypical,hu2021leveraging}, either the features extracted by the backbone $f(\mathbf{x})$ are directly used, or the backbone is freezed during fine-tuning. In other words, $\hat{f}(\mathbf{x}) \equiv f(\mathbf{x})$ and the capacity of the model is only determined by the linear classification head $W$ (or its alike), which has very low capacity, and in turn leads to high bias according to the bias-variance tradeoff. Other methods~\cite{dhillon2020baseline} fine-tune the entire backbone network, i.e., all parameters in the backbone $f(\mathbf{x})$ are learnable and $\hat{f}(\mathbf{x})$ is completely different from $f(\mathbf{x})$. Such a high capacity model inevitably leads to overfitting and ultra high variance, despite having a small bias. As aforementioned, we need to strike a good balance.

We propose a simple remedy called adaptability calibration (AC), which freezes \emph{part of the layers} in the backbone $f(\mathbf{x})$ and fine-tune the rest layers to form $\hat{f}(\mathbf{x})$. The less layers frozen, the higher is the model's adaptability. The calibration is experimentally determined.

\begin{figure}[t]
    \centering
    \subfigure[1-shot]{
        \includegraphics[width=0.4\linewidth]{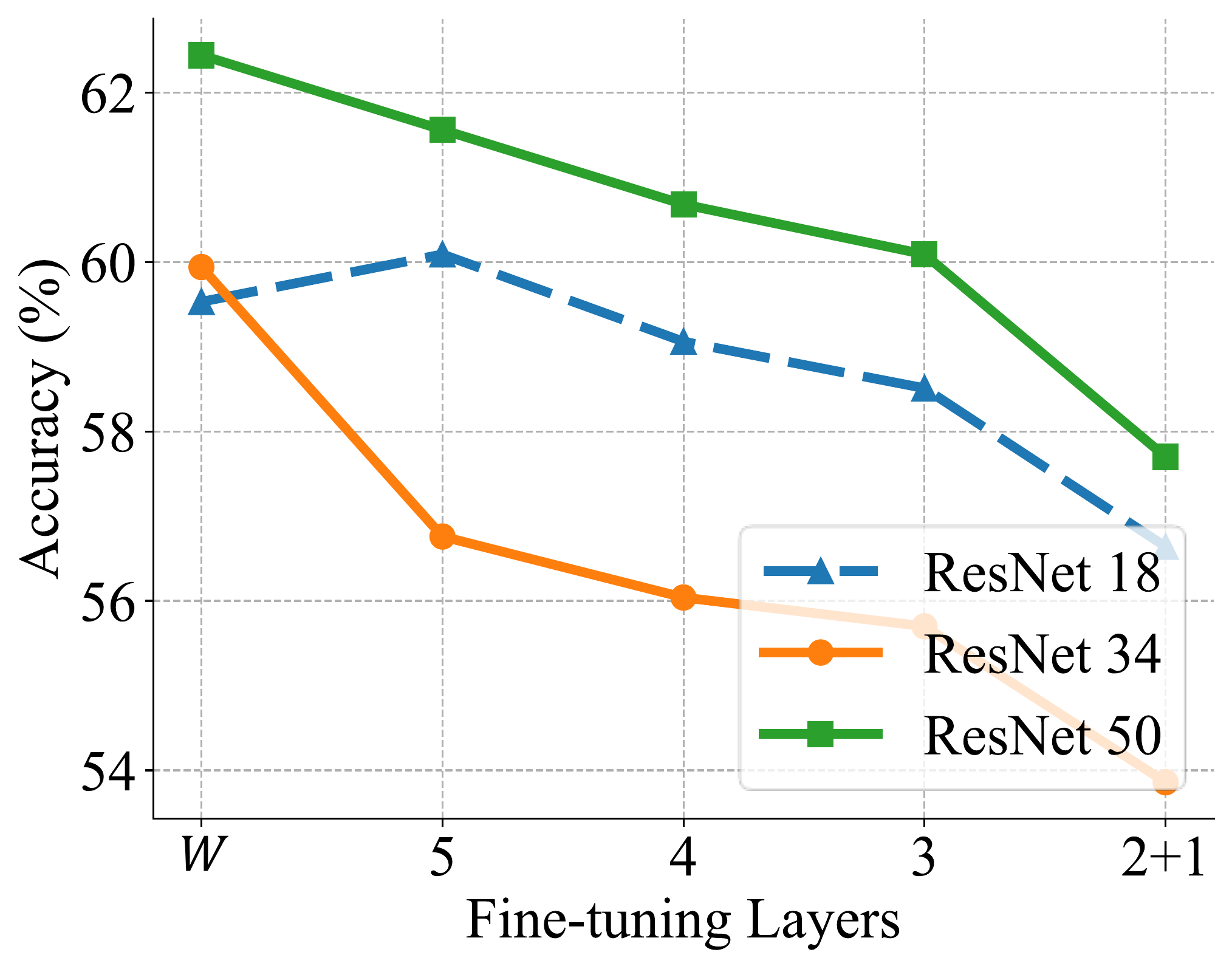}
        \label{fig:pre_k_1}
    }
    \subfigure[5-shot]{
        \includegraphics[width=0.4\linewidth]{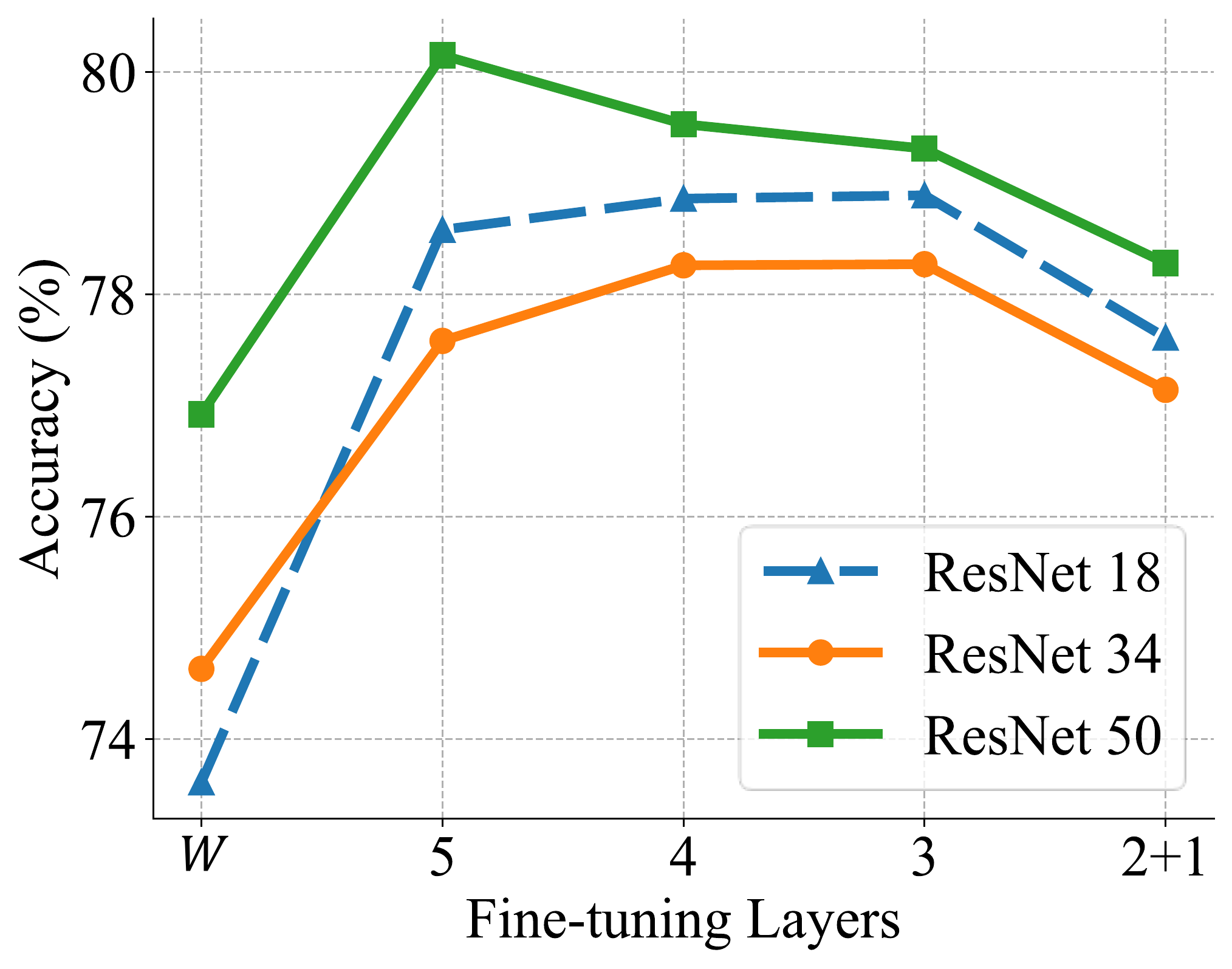}
        \label{fig:pre_k_5}
    }
	\caption{Results of 5-way accuracy on \emph{mini}-ImageNet w.r.t. the number of learnable blocks using the ResNet backbone~\cite{he2016deep}. The left-most column means only $W$ is learnable. The column `5' means $W$ plus the `res5' group in ResNet are learnable, etc. The whole backbone together with $W$ are updated at the right-most column `2+1'.}
    \label{fig:calibration}
\end{figure}

Backbone models in the ResNet~\cite{he2016deep} family have 5 groups of residual blocks, denoted as `res1' (close to the input) till `res5' (close to the classification head $W$). In Fig.~\ref{fig:calibration}, from left to right we gradually make `res5' to `res1' learnable. In the 5-shot scenario, allowing both $W$ and `res5' updatable in fact consistently strikes the best tradeoff between bias and variance. However, in the difficult 1-shot case, only learning $W$ seems to be the best while $W$+`res5' is the runner-up. Hence, considering both cases, our adaptability calibration chooses to update both $W$ and `res5' (the column `5' in Fig.~\ref{fig:calibration}).

\subsection{Reducing both Variance and Bias: Model Ensemble}

It is well-known that model ensemble is an effective way in reducing both the variance and the bias of a classifier system~\cite{zhou2012ensemble}, and hence it is potentially useful for our task. Diversity among models in the ensemble (i.e., models need to predict differently for the same input) is crucial to the success of ensemble learning, while diversity is often obtained by sampling different training sets (or example weights) for different models on the \emph{same} task (e.g., in one episode of few-shot learning).

However, the sampled novel sets for fine-tuning few-shot recognizers are so small that it leaves almost no space for sampling training sets once more for ensemble learning (e.g., consider a 1-shot task). With the proposed stability regularization and its inclusion of the base set $D_b$, ensemble learning becomes possible and natural.

We randomly divide the base set $D_b$ into $M=4$ disjoint subsets, $D_b^m$ ($m=1,2,3,4$). Then we train $M$ few-shot recognizers for \emph{one} episode, each trained with the sampled novel set and stability regularized by one $D_b^m$. During inference, the classification probabilities output by these $M$ models are averaged to generate recognition results of the ensemble model.

\section{Experiments}

\subsection{Implementation Details}

\textbf{Datasets.}
We evaluate the proposed method on three benchmark datasets, which are \textit{mini}-ImageNet~\cite{vinyals2016matching}, CUB-200-2011 (CUB)~\cite{wah2011caltech} and CIFAR-FS~\cite{bertinetto2018meta}. \textit{mini}-ImageNet consists of 100 categories randomly selected from the ImageNet dataset~\cite{russakovsky2015imagenet} with each category containing 600 images sized 84$\times$84. Following~\cite{ravi2016optimization}, it is further split to 64 base, 16 validation and 20 novel categories. CUB is composed of 200 classes (11,788 images) with size of 84$\times$84, which is split into 100 base, 50 validation and 50 novel categories. CIFAR-FS is produced by arbitrarily dividing CIFAR-100~\cite{cifar_100} into 64 base, 16 validation and 20 novel classes. There are 600 images with size of 32$\times$32 pixels within each class.

\textbf{Training Details.}
To validate the effectiveness of the proposed method, we take WRN-28-10~\cite{zagoruyko2016wide} or models in the ResNet series~\cite{he2016deep} as the backbone $f(\mathbf{x})$, and the cosine classifier used in~\cite{chen2019closer} as the classification head $W$. We follow~\cite{mangla2020charting} to conduct pretraining. At the fine-tuning stage, we train $\hat{f}(\mathbf{x})$ and $W$ by adapting the pretrained model $f(\mathbf{x})$ to novel categories. For each few-shot episode, the sampled novel set is used to fine-tune the model for 100 epochs with the label smoothed cross-entropy loss~\cite{szegedy2016rethinking} ($\epsilon$=0.1), by using SGD with learning rate 0.1, weight decay 1e-4 and momentum 0.9. 

\textbf{Evaluation Protocols.} We conduct evaluation in the form of $N$-way $K$-shot $Q$-query, where firstly $N$ categories from the novel split of different datasets are selected arbitrarily, then mutually exclusive $K$ and $Q$ samples of each category are further chosen in the same manner, respectively. The $N\times K$ samples are regarded as $D_n$ to adapt the pretrained model to these new categories, afterwards the model is evaluated on the $N\times Q$ test samples. The above process is collectively called one episode. We keep $N$=5, $Q$=15 and vary $K$=1 or 5 to conduct experiments. We report the average of all ($\text{Acc}_m$), worst 1 ($\text{Acc}_1$), average of worst 10 ($\text{Acc}_{10}$), average of worst 100 ($\text{Acc}_{100}$) accuracy and standard deviation ($\sigma$) over 500 pre-sampled episodes by 5 runs then average. Specifically for previous methods with only $\text{Acc}_m$ reported, we get other results following their official implementations. All experiments we conduct on one dataset are based on the same 500 episodes, which makes the comparison fair.

\subsection{Comparing with State-of-the-art Methods}

We compare 5-way accuracy of our method with state-of-the-art methods in Table~\ref{tab:worst_case}. `AC' and `SR' stand for adaptability calibration and stability regularization, respectively, and `EnSR' means ensemble of SR. 

\begin{table*}[t]
\centering
\caption{5-way accuracy~(\%) on \textit{mini}-ImageNet, CUB and CIFAR-FS with WRN-28-10 as the backbone. $\text{ACC}_m$ is the average accuracy copied from published papers if not otherwise noted.} 
\resizebox{.9\linewidth}{!}{
    \begin{tabular}{rc|l|ccccccc|ccccccc}
        \toprule
        & \multirow{2}*{Dataset} & \multirow{2}*{Method}
        &\multicolumn{7}{c|}{\textbf{1-shot}}
        &\multicolumn{7}{c}{\textbf{5-shot}} \\
        & & & $\text{ACC}_m$ && $\text{ACC}_1$ & & $\text{ACC}_{10}$ && $\text{ACC}_{100}$ & $\text{ACC}_m$ && $\text{ACC}_1$ && $\text{ACC}_{10}$ && $\text{ACC}_{100}$ \\ 
        \midrule
        &\multirow{11}*{\textit{mini}-ImageNet}  
        & ProtoNet$^\dagger$~\cite{snell2017prototypical}       & 54.16 && 19.76 && 26.08 && 37.62 & 73.68 && 43.74 && 49.78 && 59.46 \\
        & & Negative-Cosine~\cite{liu2020negative}    & 61.72 && 24.27 && 36.13 && 46.92 & 81.79 && 53.30 && 58.12 && 68.86 \\
        & & MixtFSL~\cite{afrasiyabi2021mixture}      & 64.31 && 30.67 && 35.07 && 46.68 & 81.66 && 46.67 && 60.13 && 71.23 \\
        & & $\text{S2M2}_R$~\cite{mangla2020charting} & 64.93 && 37.58 && 42.87 && 53.40 & 83.18 && 58.66 && 66.21 && 74.73 \\
        & & PT+NCM~\cite{hu2021leveraging}            & 65.35 && 32.00 && 38.13 && 48.41 & 83.87 && 56.00 && 64.00 && 73.89 \\
        & & Transductive-FT~\cite{dhillon2020baseline}& 65.73 && 24.00 && 33.73 && 47.60 & 78.40 && 50.67 && 53.73 && 63.09 \\
        & & CGCS~\cite{Gao_2021_ICCV}                 & 67.02 && 38.70 && 44.00 && 53.50 & 82.32 && 49.30 && 56.30 && 67.30 \\
        & & LR-DC~\cite{yang2021free}                 & 68.57 && 37.33 && 42.72 && 53.54 & 82.88 && 60.52 && 64.98 && 74.24 \\
        \cline{3-17}
        & & AC+SR (ours)                              & 69.38 && 40.52 && 44.51 && 54.97 & 85.87 && 63.20 && 66.51 && 76.28 \\
        & & AC+EnSR (ours)                            & \textbf{69.59} && \textbf{40.52} && \textbf{44.94} && \textbf{55.32} & \textbf{85.97} && \textbf{63.48} && \textbf{66.74} && \textbf{76.40} \\
        \midrule
        \midrule
        &\multirow{10}*{CUB}  
        &   Negative-Cosine~\cite{liu2020negative}    & 72.66 && 36.00 && 42.20 && 55.22 & 89.40 && 70.70 && 72.69 && 79.51 \\
        & & ProtoNet$^\dagger$~\cite{snell2017prototypical}   & 72.99 && 28.00 && 35.14 && 47.90 & 86.64 && 53.33 && 60.65 && 70.20 \\
        & & MixtFSL~\cite{afrasiyabi2021mixture}      & 73.94 && 40.00 && 44.93 && 55.76 & 86.01 && 57.33 && 66.40 && 76.65 \\
        & & CGCS~\cite{Gao_2021_ICCV}                 & 74.66 && 50.67 && 56.00 && 68.59 & 88.37 && 57.33 && 63.33 && 72.76 \\
        & & LR-DC~\cite{yang2021free}                 & 79.56 && 44.00 && 54.80 && 66.52 & 90.67 && 68.80 && 76.16 && 84.30 \\
        & & PT+NCM~\cite{hu2021leveraging}            & 80.57 && 40.00 && 52.67 && 65.83 & 91.15 && 69.33 && 76.13 && 84.67 \\
        & & $\text{S2M2}_R$~\cite{mangla2020charting} & 80.68 && 52.00 && 56.56 && 69.70 & 90.85 && 73.86 && 77.72 && 85.47 \\
        \cline{3-17}
        & & AC+SR (ours)                              & 85.14 && 52.78 && 57.46 && 71.31 & 94.42 && 76.00 && 80.83 && 87.76 \\
        & & AC+EnSR (ours)                            & \textbf{85.42} && \textbf{53.04} && \textbf{58.10} && \textbf{71.58} & \textbf{94.53} && \textbf{77.58}  && \textbf{81.12} && \textbf{87.99} \\
        \midrule
        \midrule
        &\multirow{11}*{CIFAR-FS}  
        & ProtoNet$^{\Diamond}$~\cite{snell2017prototypical}       & 61.60 && 25.34 && 30.62 && 43.08 & 79.08 && 51.20 && 57.40 && 66.62 \\
        & & Inductive-FT~\cite{dhillon2020baseline}   & 68.72 && 33.33 && 38.40 && 50.32 & 86.11 && 56.00 && 61.47 && 71.73 \\
        & & Negative-Cosine$^\Diamond$~\cite{liu2020negative}    & 68.90 && 26.66 && 39.76 && 52.58 & 83.82 && 52.00 && 61.54 && 71.88 \\
        & & MixtFSL$^\Diamond$~\cite{afrasiyabi2021mixture}      & 69.42 && 29.33 && 38.53 && 51.51 & 81.05 && 57.33 && 62.40 && 69.19 \\
        & & LR-DC$^\Diamond$~\cite{yang2021free}                 & 72.52 && 32.26 && 42.08 && 54.72 & 83.92 && 57.60 && 63.30 && 72.48 \\
        & & CGCS~\cite{Gao_2021_ICCV}                 & 73.00 && 33.33 && 41.33 && 55.53 & 85.80 && 56.00 && 62.13 && 69.16 \\
        & & PT+NCM~\cite{hu2021leveraging}            & 74.64 && 34.67 && 42.53 && 56.60 & \textbf{87.64} && 54.67 && 62.13 && 71.57 \\
        & & $\text{S2M2}_R$~\cite{mangla2020charting} & 74.81 && 35.74 && 42.06 && 56.49 & 87.47 && 56.26 && 63.38 && 72.17 \\
        \cline{3-17}
        & & AC+SR (ours)                              & 74.00 && \textbf{36.54} && 42.73 && 56.83 & 86.65 && 56.26 && 65.97 && 75.23 \\
        & & AC+EnSR (ours)                            & \textbf{74.85} && 34.96 && \textbf{42.91} && \textbf{57.70} & 87.27 && \textbf{59.48} && \textbf{66.61} && \textbf{75.84} \\
        \bottomrule
    \end{tabular}} \\
    $^{\Diamond}$ $\text{ACC}_m$ from our implementation \quad   $^{\dagger}$ $\text{ACC}_m$ from~\cite{chen2019closer}
    \label{tab:worst_case}
\end{table*}

The proposed method (denoted as AC+SR) outperforms existing methods using the traditional $\text{ACC}_m$ metric by 1--5\% in both 1-shot and 5-shot scenarios on the first two datasets, and is slightly ($<1\%$) worse on CIFAR-FS. When model ensemble is used, our AC+EnSR almost consistently outperforms existing methods.

More importantly, in the worst-case ($\text{ACC}_1$) and near-worst case average metrics ($\text{ACC}_{10}$, $\text{ACC}_{100}$), our methods are almost consistently better, and mostly higher by $>2\%$ margins. In the worst-case metrics, model ensemble (AC+EnSR) is again better than single model. And, the ranking from $\text{ACC}_m$ is significantly different from that based on $\text{ACC}_1$, but the ranking orders are almost consistent among the 3 worst-case metrics. Hence, the worst-case metric is more \emph{stable} than the average accuracy metric.

\subsection{Relationships: $\mu$, $\sigma$, $\mu-3\sigma$, $\text{ACC}_1$ and $\text{ACC}_m$}

Table~\ref{tab:lower} presents a few important evaluation metrics, including the commonly used average accuracy $\text{ACC}_m$ ($\mu$), the proposed worst-case accuracy $\text{ACC}_1$ that we advocate, the standard deviation $\sigma$, and the surrogate $\mu-3\sigma$ that motivates the proposed method.

\begin{table*}[t]
\centering
\caption{Key evaluation metrics, including the commonly used average accuracy $\text{ACC}_m$ ($\mu$), the proposed worst-case accuracy $\text{ACC}_1$, the standard deviation $\sigma$, and the surrogate $\mu-3\sigma$ that motivates the proposed method.}
\resizebox{.9\linewidth}{!}{
    \begin{tabular}{rc|l|ccccccc|ccccccc}
        \toprule
        & \multirow{2}*{Dataset} & \multirow{2}*{Method}
        &\multicolumn{7}{c|}{\textbf{1-shot}}
        &\multicolumn{7}{c}{\textbf{5-shot}} \\
        & & & $\text{ACC}_1$ && $\mu-3\sigma$ && $\sigma$ && $\text{ACC}_m$ & $\text{ACC}_1$ && $\mu-3\sigma$ && $\sigma$ && $\text{ACC}_m$ \\ 
        \midrule
        &\multirow{11}*{\textit{mini}-ImageNet}  
        & ProtoNet$^\dagger$~\cite{snell2017prototypical}      & 19.76 && 23.41 && 10.25 && 54.16 & 43.74 && 49.32 && 8.12 && 73.68\\
        & & Transductive-FT~\cite{dhillon2020baseline}& 24.00 && 32.82 && 10.97 && 65.73 & 50.67 && 53.23 && 8.39 && 78.40 \\
        & & Negative-Cosine~\cite{liu2020negative}    & 24.27 && 31.36 && 10.12 && 61.72 & 53.30 && 61.18 && 6.87 && 81.79 \\
        & & MixtFSL~\cite{afrasiyabi2021mixture}      & 30.67 && 34.70 &&\, 9.87 && 64.31 & 46.67 && 59.16 && 7.50 && 81.66   \\
        & & PT+NCM~\cite{hu2021leveraging}            & 32.00 && 34.75 && 10.20 && 65.35 & 56.00 && 63.98 && 6.63 && 83.87 \\
        & & LR-DC~\cite{yang2021free}                 & 37.33 && 37.73 && 10.28 && 68.57 & 60.52 && 63.02 && 6.62 && 82.88 \\
        & & $\text{S2M2}_R$~\cite{mangla2020charting} & 37.58 && 37.39 &&\, \textbf{9.18} && 64.93 & 58.66 && 66.35 && \textbf{5.61} && 83.18 \\
        & & CGCS~\cite{Gao_2021_ICCV}                 & 38.70 && 36.42 && 10.20 && 67.02 & 49.30 && 60.90 && 7.14 && 82.32 \\
        \cline{3-17}
        & & AC+SR (ours)                              & 40.52 && 40.25 &&\, 9.71 && 69.38 & 63.20 && 66.46 && 6.47 && 85.87  \\
        & & AC+EnSR (ours)                            & \textbf{40.52} && \textbf{40.67} &&\, 9.64 && \textbf{69.59} & \textbf{63.48} && \textbf{66.71} && 6.42 && \textbf{85.97} \\
        \midrule
        \midrule
        &\multirow{10}*{CUB}  
        &   ProtoNet$^\dagger$~\cite{snell2017prototypical}   & 28.00 && 39.99 && 11.00 && 72.99 & 53.33 && 67.53 && 6.37 && 86.64 \\
        & & Negative-Cosine~\cite{liu2020negative}            & 36.00 && 40.80 && 10.62 && 72.66 & 70.70 && 73.29 && 5.37 && 89.40 \\
        & & PT+NCM~\cite{hu2021leveraging}                    & 40.00 && 49.97 && 10.20 && 80.57 & 69.33 && 75.85 && 5.10 && 91.15 \\
        & & MixtFSL~\cite{afrasiyabi2021mixture}              & 40.00 && 32.69 && 13.75 && 73.94 & 57.33 && 67.26 && 6.25 && 86.01 \\
        & & LR-DC~\cite{yang2021free}                         & 44.00 && 49.74 &&\, 9.94 && 79.56 & 68.80 && 75.49 && 5.06 && 90.67 \\
        & & CGCS~\cite{Gao_2021_ICCV}                         & 50.67 && 42.53 && 10.71 && 74.66 & 57.33 && 70.01 && 6.12 && 88.37 \\
        & & $\text{S2M2}_R$~\cite{mangla2020charting}         & 52.00 && 50.32 && 10.12 && 80.68 & 73.86 && 74.35 && 5.50 && 90.85 \\
        \cline{3-17}
        & & AC+SR (ours)                              & 52.78 && 58.44 && 8.90 && 85.14 & 76.00 && 81.85 && 4.19 && 94.42 \\
        & & AC+EnSR (ours)                            & \textbf{53.04} && \textbf{58.84} && \textbf{8.86} && \textbf{85.42} & \textbf{77.58} && \textbf{82.14} && \textbf{4.13} && \textbf{94.53} \\
        \midrule
        \midrule
        &\multirow{11}*{CIFAR-FS}  
        & ProtoNet$^{\Diamond}$~\cite{snell2017prototypical}     & 25.34 && 23.38 && 12.74 && 61.60 & 51.20 && 53.58 && 8.50 && 79.08 \\
        & & Negative-Cosine$^\Diamond$~\cite{liu2020negative}    & 26.66 && 34.49 && 11.47 && 68.90 & 52.00 && 60.21 && 7.87 && 83.82 \\
        & & MixtFSL$^\Diamond$~\cite{afrasiyabi2021mixture}      & 29.33 && 33.09 && 12.11 && 69.42 & 57.33 && 56.48 && 8.19 && 81.05 \\
        & & LR-DC$^\Diamond$~\cite{yang2021free}                 & 32.26 && 36.94 && 11.86 && 72.52 & 57.60 && 60.94 && 7.66 && 83.92 \\
        & & Inductive-FT~\cite{dhillon2020baseline}              & 33.33 && 36.29 && 10.81 && 68.72 & 56.00 && 63.37 && 7.58 && 86.11 \\
        & & CGCS~\cite{Gao_2021_ICCV}                            & 33.33 && 39.13 && 11.29 && 73.00 & 56.00 && 61.59 && 8.07 && 85.80 \\
        & & PT+NCM~\cite{hu2021leveraging}                       & 34.67 && 42.51 && 10.71 && 74.64 & 54.67 && 64.69 && 7.65 && \textbf{87.64} \\
        & & $\text{S2M2}_R$~\cite{mangla2020charting}            & 35.74 && \textbf{45.74} &&\, \textbf{9.69} && 74.81 & 56.26 && \textbf{67.58} && \textbf{6.63} && 87.47 \\
        \cline{3-17}
        & & AC+SR (ours)                              & \textbf{36.54} && 39.35 && 11.55 && 74.00 & 56.26 && 64.45 && 7.40 && 86.65 \\
        & & AC+EnSR (ours)                            & 34.96 && 40.14 && 11.57 && \textbf{74.85} & \textbf{59.48} && 65.16 && 7.37 && 87.27 \\
        \bottomrule
    \end{tabular}} \\
    $^{\Diamond}$ $\text{ACC}_m$ from our implementation \quad   $^{\dagger}$ $\text{ACC}_m$ from~\cite{chen2019closer}
    \label{tab:lower}
\end{table*}

As aforementioned, because the worst-case accuracy $\text{ACC}_1$ is unable to be directly modeled or optimized using few-shot samples, we alternatively resort to the surrogate $\mu-3\sigma$ inspired by the $3\sigma$ rule. This objective is again difficult because $\sigma$ is unable to be directly accessed, and we resort to the bias-variance decomposition: reducing $\sigma$ and increasing $\mu$ simultaneously.

As Table~\ref{tab:lower} shows, both $\mu$ and $\sigma$ are correlated with $\text{ACC}_1$, but not tightly correlated. For example, $\text{S2M2}_R$~\cite{mangla2020charting} often has the smallest $\sigma$, but because its $\mu$ ($\text{ACC}_m$) is not the most competitive, its worst-case $\text{ACC}_1$ is significantly lower than ours. We do need to optimize both $\mu$ and $\sigma$ simultaneously.

$\mu-3\sigma$, which motivates our method, correlates better with $\text{ACC}_1$ than $\sigma$ or $\mu$, as Table~\ref{tab:lower} shows. However, this quantity is still far from being a perfect surrogate. It is still a great challenge to model the worst-case $\text{ACC}_1$ end-to-end.

\subsection{Ablation Analyses}
\label{sec:ablation}

\textbf{Switching backbone.}
Table~\ref{tab:backbone} shows the results of 5-way classification on \textit{mini}-ImageNet by switching to different backbones. Our adaptability calibration (AC) alone is sometimes harmful in the 1-shot scenario, but always beneficial for 5-shot, which coincides well with the observations in Fig.~\ref{fig:calibration}. 

\begin{table*}[t]
\footnotesize
\centering
\caption{Ablation results of 5-way classification on \emph{mini}-ImageNet with different backbones. The first row is the baseline and our AC or SR are gradually incorporated. Note that EnSR always includes SR.} 
\resizebox{.8\linewidth}{!}{
    \begin{tabular}{rc|ccc|ccccccccc|ccccccccc}
        \toprule
        & \multirow{2}*{Backbone} & \multirow{2}*{AC} & \multirow{2}*{SR} & \multirow{2}*{EnSR}
        &\multicolumn{9}{c|}{\textbf{1-shot}}
        &\multicolumn{9}{c}{\textbf{5-shot}} \\
        & & & & & $\text{ACC}_{m}$ && $\sigma$ && $\text{ACC}_1$ && $\text{ACC}_{10}$ && $\text{ACC}_{100}$ & $\text{ACC}_m$  && $\sigma$ && $\text{ACC}_1$ && $\text{ACC}_{10}$ && $\text{ACC}_{100}$ \\
        \midrule
        &\multirow{4}*{ResNet 18}  
        & \ & \ & \   & 59.53 && \textbf{9.99}  && 30.42 && 35.09 && 45.05 & 73.61 && 8.16 && 44.54 && 50.16 && 61.36 \\
        & & \Checkmark & \ & \   & 60.09 && 10.13 && 25.60 && 34.55 && 45.24 & 78.58 && 7.62 && 45.86 && 56.34 && 67.32 \\
        & & \Checkmark & \Checkmark & \     & 62.33 && 10.27	&& 32.00 &&	37.03 && 47.36 & 79.02 && 7.68 && 46.14 && 56.48 && 67.62 \\
        & &  \Checkmark & \Checkmark & \Checkmark    & \textbf{62.76} && 10.16 && \textbf{32.78} && \textbf{37.55} && \textbf{47.84} & \textbf{79.23} && \textbf{7.62} && \textbf{46.14} && \textbf{56.96} && \textbf{67.91} \\
        \midrule
        \midrule
        &\multirow{4}*{ResNet 34}  
        & \ & \ & \   & 59.94 && \textbf{9.64} && 30.94 && 35.73 && 46.08 & 74.63 && 7.68 && 45.60 && 52.25 && 63.17 \\
        & & \Checkmark & \ & \   & 56.76 && 9.81 && 24.56 && 32.09 && 42.51 & 77.58 && 7.54 && 47.48 && 55.54 && 66.38 \\
        & & \Checkmark & \Checkmark & \     & 61.42 && 9.87 && 32.54 && 37.17 && 47.00 & 78.59 && 7.43 && 51.46 && 57.76 && 67.43 \\
        & & \Checkmark & \Checkmark & \Checkmark     & \textbf{61.80} && 9.90 && \textbf{33.86} && \textbf{37.76} && \textbf{47.26} & \textbf{78.74} && \textbf{7.41} && \textbf{51.74} && \textbf{58.03} && \textbf{67.58} \\
        \midrule
        \midrule
        &\multirow{4}*{ResNet 50}  
        & \ & \ & \   & 62.44 && 10.04 && 29.32 && 36.64 && 47.74 & 76.92 && 7.93 && 50.66 && 54.67 &&	65.04 \\
        & & \Checkmark & \ & \   & 61.56 && 9.85  && 29.06 && 36.69 && 47.25 & 80.15 && \textbf{7.21} && 52.26 && \textbf{59.07} && \textbf{69.29} \\
        & & \Checkmark & \Checkmark & \     & 63.60 && 9.90  && 30.94 &&	37.36 && 49.00 & 80.33 && 7.46 && 53.06 && 57.89 && 68.93 \\
        & & \Checkmark & \Checkmark & \Checkmark     & \textbf{63.79} && \textbf{9.84}  && \textbf{31.67} && \textbf{38.05} && \textbf{49.36} & \textbf{80.43} && 7.36 && \textbf{53.30} && 58.89 && 69.24 \\
        \midrule
        \midrule
         &\multirow{4}*{WRN-28-10}  
        & \ & \ & \   & 67.86 && 9.89 && 37.58 && 42.87 && 53.40 & 84.57 && 6.50 && 58.66 && 66.21 && 74.73 \\
        & & \Checkmark & \ & \   & 67.36 && 9.79 && 36.80 && 41.44 && 52.80 & 85.53 && 6.50 && 60.54 && 65.63 && 75.95 \\
        & & \Checkmark & \Checkmark & \   & 69.38 && 9.71 && 40.52 && 44.51 && 54.97 & 85.87 && 6.47 && 63.20 && 66.51 && 76.28 \\
        & & \Checkmark & \Checkmark & \Checkmark & \textbf{69.59} && \textbf{9.64} && \textbf{40.52} && \textbf{44.94} && \textbf{55.32} & \textbf{85.97} && \textbf{6.42} && \textbf{63.48} && \textbf{66.74} && \textbf{76.40} \\
        \bottomrule
    \end{tabular}}
    \label{tab:backbone}
\end{table*}

On the other hand, the proposed stability regularization (SR) is consistently helpful in all cases and all metrics, especially in the 1-shot case. The proposed model ensemble method (EnSR) is almost consistently improving all evaluation metrics on all backbone models, in particular for the difficult 1-shot scenario.

\textbf{Switching pretraining methods.}
We further conduct experiments to explore the influence of pretraining methods on the proposed strategies. Specifically, we additionally adopt two pretraining methods from~\cite{chen2019closer}, denoted as Baseline and Baseline++, respectively, for simplicity. As shown in Table~\ref{tab:method}, when pretrained with Baseline, after applying AC and SR, $\text{ACC}_1$ and $\text{ACC}_{10}$ are increased (on average by 2\% in 1-shot and 1\% in 5-shot), although $\text{ACC}_m$ is slightly decreased (about 0.2\%). This observation not only shows our AC+SR's effectiveness, but also indicates that average and worst-case accuracy are not aligned in general. With AC+EnSR, the accuracy for all metrics gains significant margins than the naive Baseline method, in particular, 4-5\% for 1-shot.

\begin{table*}[t]
\footnotesize
\centering
\caption{Ablation results of 5-way classification on \textit{mini}-ImageNet with the ResNet 18 backbone pretrained by other methods.} 
\resizebox{.9\linewidth}{!}{
    \begin{tabular}{rc|ccc|ccccccccc|ccccccccc}
        \toprule
        & \multirow{2}*{Pretraining Method} & \multirow{2}*{AC} & \multirow{2}*{SR} & \multirow{2}*{EnSR}
        &\multicolumn{9}{c|}{\textbf{1-shot}}
        &\multicolumn{9}{c}{\textbf{5-shot}} \\
        & & & & & $\text{ACC}_{m}$ && $\sigma$ && $\text{ACC}_1$ && $\text{ACC}_{10}$ && $\text{ACC}_{100}$ & $\text{ACC}_m$  && $\sigma$ && $\text{ACC}_1$ && $\text{ACC}_{10}$ && $\text{ACC}_{100}$ \\
        \midrule
        &\multirow{3}*{Baseline~\cite{chen2019closer}}  
        & \ & \ & \   & 50.37 && 10.59 && 18.42 && 24.91 && 35.43 & 74.13 && \textbf{8.14} && 44.82 && 51.20 && 62.15 \\
        & & \Checkmark & \Checkmark & \     & 50.27 && 10.37 && 21.86 && 26.45 && 35.43 & 73.92 && 8.22 && 46.94 && 51.49 && 61.85 \\
        & & \Checkmark & \Checkmark & \Checkmark     & \textbf{54.42} && \textbf{10.26} && \textbf{23.20} && \textbf{29.39} && \textbf{39.61} & \textbf{74.88} && 8.16 && \textbf{47.46} && \textbf{52.89} && \textbf{62.78} \\
        \midrule
        \midrule
        &\multirow{3}*{Baseline++~\cite{chen2019closer}}  
        & \ & \ & \   & 57.59 && \textbf{9.50} && 29.06 && 33.58 && 43.99 & 75.35 && 7.66 && 49.08 && 55.66 && 64.17 \\
        & & \Checkmark & \Checkmark & \     & 59.30 && 9.91 && 29.58 && 34.08 && 44.77 & 77.57 && 7.63 && 54.68 && 58.36 && 66.19 \\
        & & \Checkmark & \Checkmark & \Checkmark     & \textbf{59.58} && 9.82 && \textbf{30.66} && \textbf{34.83} && \textbf{45.20} & \textbf{77.73} && \textbf{7.60} && \textbf{56.80} && \textbf{58.94} && \textbf{66.51} \\
        \bottomrule
    \end{tabular}}
    \label{tab:method}
\end{table*}

When pretraining with Baseline++, the accuracy in all cases are generally improved by gradually deploying AC+SR or AC+EnSR, too. The proposed techniques are compatible with various pretraining methods.

\textbf{Calibrating adaptability differently.}
Table~\ref{tab:AC_degree} shows the results by varying the number of learnable convolutional groups in ResNet 18 to experiment with different calibration of adaptability. First, there is a clear distinction between the first row ($W$ only) and other rows, indicating that it is harmful to freeze the entire backbone and only fine-tune the classification head $W$.

\begin{table*}[t]
\footnotesize
\centering
\caption{Ablation results of 5-way classification on \textit{mini}-ImageNet with ResNet 18 as the backbone by varying the number of learnable blocks (\Checkmark means learnable) to control the degree of AC. The number beneath the column AC stands for the index of convolutional groups in ResNet 18. AC+EnSR is used in all experiments.} 
\resizebox{.8\linewidth}{!}{
    \begin{tabular}{rccccc|ccccccccc|ccccccccc}
        \toprule
        & \multirow{2}*{$W$} & \multicolumn{4}{c|}{AC}
        &\multicolumn{9}{c|}{\textbf{1-shot}}
        &\multicolumn{9}{c}{\textbf{5-shot}} \\
        & & 5 & 4 & 3 & 2+1 & $\text{ACC}_{m}$ && $\sigma$ && $\text{ACC}_1$ && $\text{ACC}_{10}$ && $\text{ACC}_{100}$ & $\text{ACC}_m$  && $\sigma$ && $\text{ACC}_1$ && $\text{ACC}_{10}$ && $\text{ACC}_{100}$ \\
        \midrule
        & \Checkmark & & & & & 59.53 && \textbf{9.99} && 30.42 && 35.09 && 45.05 & 73.61 && 8.16 && 44.54 && 50.16 && 61.36 \\
        & \Checkmark & \Checkmark & & & & 62.76 && 10.16 && \textbf{32.78} && \textbf{37.55} && \textbf{47.84} & 79.23 && 7.62 && 46.14 && 56.96 && 67.91 \\
        & \Checkmark & \Checkmark & \Checkmark & & & 62.77 && 10.38 && 30.12 && 35.44 && 47.46 & 79.91 && \textbf{7.54} && 45.60 && \textbf{57.84} &&	68.77 \\
        & \Checkmark & \Checkmark & \Checkmark & \Checkmark & & \textbf{62.80} && 10.50 && 22.96 && 34.93 && 47.27 & \textbf{80.04} && 7.57 && 45.58 &&	57.74 && \textbf{68.85} \\
        & \Checkmark & \Checkmark & \Checkmark & \Checkmark & \Checkmark & 62.19 && 10.50 && 25.60 && 35.29 && 46.68 & 79.87 && 7.58 &&	\textbf{48.00} && 57.41 && 68.67 \\
        \bottomrule
    \end{tabular}}
    \label{tab:AC_degree}
\end{table*}

Second, when more residual groups are made learnable, there is not a clear winner, especially when we consider different evaluation metrics. For example, every row in Table~\ref{tab:AC_degree}'s last 3 rows has been the champion for at least one evaluation metric in 5-shot. But, if we consider 1-shot, it is obviously observed that the second row (fine-tuning $W$+`res5') is significantly better. This is the adaptability calibration we choose in our experiments.

\textbf{Generalize SR to other data.}
Since only raw images are required to compute the SR loss, it enjoys the flexibility to use others images (i.e., those not from $D_b$) for its computation (i.e., being generalizable). We use CUB~\cite{wah2011caltech}, Cars~\cite{krause20133d}, Describable Textures (DTD)~\cite{cimpoi2014describing}, Pets~\cite{parkhi2012cats}, VGG Flower (Flower)~\cite{nilsback2008automated} and CIFAR-100~\cite{cifar_100} as the images to compute our stability regularization loss when the base set is \emph{mini}-ImageNet. All these images are resized to 84$\times$84. 

Results in Table~\ref{tab:ext} reveals an interesting observation. No matter what images are used to compute the SR loss, the accuracy is consistently higher than the baseline (AC without SR), regardless of which metric is considered. That is, the stability regularization is indeed generalizable.

\begin{table*}[t]
\footnotesize
\centering
\caption{5-way 1-shot results on \textit{mini}-ImageNet by deploying AC+SR with ResNet 18 as the backbone, while the SR loss is computed with images from \emph{different} datasets, denoted as $D_{sr}$. The first row (`$-$') is the baseline with only AC, and the second row is AC+SR, where SR uses the base set (\emph{mini}-ImageNet).} 
\resizebox{.57\linewidth}{!}{
    \begin{tabular}{rl|ccccccccc}
        \toprule
        & $D_{sr}$ & $\text{ACC}_{m}$ && $\sigma$ && $\text{ACC}_1$ && $\text{ACC}_{10}$ && $\text{ACC}_{100}$ \\
        \midrule
        & - & 60.09 && 10.13 && 25.60 && 34.55 && 45.24 \\
        & \textit{mini}-ImageNet & \textbf{62.33} && 10.27 && 32.00 && \textbf{37.03} && \textbf{47.36}\\
        \midrule
        & CUB~\cite{wah2011caltech} & 60.66 && 9.99 && 30.14 && 36.43 && 46.29\\
        & Cars~\cite{krause20133d} & 61.05 && 10.29 && 30.94 && 36.64 && 46.20 \\
        & DTD~\cite{cimpoi2014describing} & 61.99 && 10.23 && 31.20 && 36.75 && 47.23\\
        & Pets~\cite{parkhi2012cats} & 61.57 && \textbf{9.95} && 32.00 && 36.62 && 47.27\\
        & Flower~\cite{nilsback2008automated} & 61.03 && 10.07 && \textbf{32.26} && 36.51 && 46.37 \\
        & CIFAR-100~\cite{cifar_100} & 60.47 && 10.04 && 30.68 && 36.69 && 46.02 \\
        \bottomrule
    \end{tabular}}
    \label{tab:ext}
\end{table*}

Moreover, although $D_b$ (the base set) performs the best in general, using other images for SR computation leads to results that are on par with it. That is, we \emph{do not} require images used in this regularization to be visually similar or semantically correlated. Our stability regularization is \emph{consistently useful}.

\section{Conclusions}

This paper advocated to use the worst-case accuracy in optimizing and evaluating few-shot learning methods, which is a better fit to real-world applications than the commonly used average accuracy. Worst-case accuracy, however, is much more difficult to work on. We designed a surrogate loss inspired by the $3\sigma$ rule, and in turn proposed two strategies to implicitly optimize this surrogate: a stability regularization (SR) loss together with model ensemble to reduce the variance, and an adaptability calibration (AC) to vary the number of learnable parameters to reduce the bias. 

The proposed strategies have achieved significantly higher worst-case (and also average) accuracy than existing methods. In the future, we will design more direct attacks to reduce the worst-case error, because the current surrogate is not a highly accurate approximation of the worst-case performance yet.

\vfill
\noindent\textbf{Acknowledgments.}  This research was partly supported by the National Natural Science Foundation of China under Grant 61921006 and Grant 61772256.

\clearpage

\bibliographystyle{splncs04}
\bibliography{egbib}

\end{document}